\documentclass[runningheads,natbib]{llncs}

\usepackage{graphicx, times}

\usepackage{wrapfig}
\usepackage{hyperref}

\hypersetup{
  colorlinks   = true, 
  urlcolor     = blue, 
  linkcolor    = blue, 
  citecolor   = blue 
}

\usepackage{amsmath,amssymb,amsfonts}
\usepackage{xcolor}
\usepackage{url}
\usepackage{latexsym}
\usepackage{booktabs}
\usepackage{graphicx}
\usepackage{multirow}

\usepackage{float}
\floatstyle{boxed} 
\restylefloat{figure}

\usepackage[square,numbers]{natbib}

\usepackage{enumitem}
\setlist{topsep=0pt, leftmargin=*}

\newcommand{\comment}[1]{}

\begin{document}

\title{Why Did You Not Compare With That?\\
Identifying Papers for Use as Baselines}
\titlerunning{Identifying Papers for Use as Baselines}

\author{Manjot Bedi\inst{1} \and
    Tanisha Pandey\inst{1} \and
    Sumit Bhatia\inst{2} \and
    Tanmoy Chakraborty\inst{1}}

\authorrunning{Bedi et al.}

\institute{IIIT Delhi, New Delhi, India \\
        \email{\{manjotb, tanisha17116, tanmoy\}@iiitd.ac.in} \and
    Media and Data Science Research Lab, Adobe Systems, India\\
        \email{Sumit.Bhatia@adobe.com}}

\maketitle 

\begin{abstract}
We propose the task of automatically identifying papers used as baselines in a scientific article. We frame the problem as a binary classification task where all the references in a paper are to be classified as either baselines or non-baselines. This is a challenging problem due to the numerous ways in which a baseline reference can appear in a paper. We develop a dataset of $2,075$ papers from ACL anthology corpus with all their references manually annotated as one of the two classes. We develop a multi-module attention-based neural classifier for the baseline classification task that outperforms four state-of-the-art citation role classification methods when applied to the baseline classification task. We also present an analysis of the errors made by the proposed classifier, eliciting the challenges that make baseline identification a challenging problem. 

\end{abstract}

\section{Introduction}
One of the common criticisms received by the authors of a scientific article during the paper review is that the method proposed in the submitted paper has not been compared with appropriate baselines. The reviewers often suggest a list of existing papers which, according to them, should have been used as baselines by the submitted work.  Oftentimes, the authors find the suggestions unexpected and surprising as they have never encountered these papers before. The reasons behind the lack of awareness of the state-of-the-art of a specific research area are two-fold -- (i) the authors have not done due diligence to explore the field completely, and/or (ii) due to the exponential growth of the number of papers published per year, many relevant papers get unnoticed. Both these problems can be addressed if we have a recommendation system that collects all the papers published in a certain field, analyzes them, and recommends a set of selected papers for a given topic/task that needs to be considered for the purpose of comparison. The current work is the first step towards the goal of building such an {\em intelligent baseline recommendation system} that can assist the authors to find and select suitable baselines for their work.

With the availability of online tools such as CiteSeerX~\cite{wu2015citeseerx}, Google Scholar~\cite{jacso2005google}, and Semantic Scholar~\cite{fricke2018semantic},  it has become convenient for researchers to search for related articles. However, these search engines provide flat recommendations and do not distinguish between the recommended papers based on how and why the recommendations are relevant to the query. For example, if the query is `citation classification models', how do we know, among the set of recommendations returned by the search engines, which one would be used to understand the {\em background} of the area, which one to explore to know the {\em datasets} used in the past to address the problem, which one to use for the purpose of {\em comparison}, etc. In short, the existing systems do not provide {\em faceted recommendations} where a facet can determine the role of a recommendation with respect to the query.

In order to build an intelligent baseline recommendation system, the first requirement is the capability to automatically identify the references in a given paper used by the paper as baselines. This capability allows creating the training corpus as well as automatically process the ever-growing stream of new papers. One may think that this problem of automatic baseline identification is trivial as a baseline reference is likely to be cited in the experiment and/or the result sections of the paper; therefore, the position information of a reference may give a precise cue about its usage in the paper. Surprisingly, we observe that this assumption does not work satisfactorily -- out of $2,075$ papers we analyze in this work, the probability of a baseline citation to appear in the experiment section is $0.73$. It indicates that around $30$\% baseline references lie in some other sections of the paper. More importantly,  only $23$\% of the references placed in the experiment section are actually used as baselines in the paper. We further observe that only $7.13\%$ papers have keywords such as  `baseline', `state-of-the-art', `gold standard' present in the headings of different sections or subsections (see discussion on error analysis in Section \ref{sec:experiments} for the other challenges). These obstacles make the problem of accurately classifying references of a given paper into baselines or non-baselines non-trivial. 

The problem of \emph{baseline classification} is closely related to the task of \textit{citation role classification} studied extensively in the literature. Notable contributions include the works by Chakraborty et al.~\cite{chakraborty2016ferosa} who proposed a faceted scientific paper recommendation system by categorizing the references into four major facets;  Dong and Schäfer~\cite{dong2011ensemble} who proposed an ensemble model to figure out  different roles of references in a paper; ; Jurgens et al.~\cite{jurgens2018measuring} who unfolded the evolution of research in a scientific field by understanding why a paper is being cited; Cohan et al.~\cite{cohan2019structural} who outperformed the methods developed by Jugens et al.~\cite{jurgens2018measuring} in the task of citation role classification. (See Section \ref{sec:rw} for more details of the related literature.) However, none of these methods are explicitly developed to address the problem of baseline recommendation. Our experiments (Section~\ref{sec:experiments}) reveal that these methods do not work well to distinguish the baseline references from other references in a given paper. 

In this paper, we consider the ACL Anthology dataset, select a subset of papers  and employ human annotators to identify the references corresponding to the baselines used in the papers (Section~\ref{sec:dataset}). We present a series of issues encountered during the annotation phase that illustrate the non-trivial nature of the problem. We then develop a multi-module attention (MMA) based neural architecture to classify references into baselines and non-baselines (Section~\ref{sec:approach}). We also adopt state-of-the-art approaches for citation role classification for a fair comparison with our methods. A detailed comparative analysis shows that the neural attention  based approach outperforms others with $0.80$ F1-score. We present a thorough error analysis to understand the reasons behind the failures of the proposed models and identify challenges that need to be addressed to build better baseline identification systems (Section~\ref{sec:experiments}). The  dataset developed and code for our proposed model is available at \url{https://github.com/sumit-research/baseline-search}.

\section{Related Work}
\label{sec:rw}

\paragraph{\textbf{Understanding the Role of Citations.}}  Stevens et al.~\cite{Stevens1964CanCI} first proposed that papers are cited due to 15 different reasons. Singh et al.~\cite{10.1145/2806416.2806566} presented the role of citation context in predicting the long term impact of researchers. Pride and Knoth~\cite{pride2020authoritative}  and Teufel et al.~\cite{teufel2006automatic} attempted to classify the roles of citations. Chakraborty
and Narayanam~\cite{chakraborty2016all} and Wan and Liu~\cite{wan2014all} argued that all citations are not equally important for a citing paper, and proposed models to measure the intensity of a citation. 
Doslu and Bingol~\cite{doslu2016context} analysed the context around a citation to rank papers from similar topics.
Cohen et al.~\cite{cohen2006reducing} showed that the automatic classification of citations could be a useful tool in systematic reviews. Chakraborty et al.~\cite{chakraborty2016ferosa} presented four reasons/tags associated with citations of a given paper -- `background' (those which are important to understand the background literature of the paper), `alternative approaches' (those which deal with the similar problem as that of the paper), `methods' (those which helped in designing the model in the paper) and `comparison' (those with which the paper is compared). Therefore, one can  simply assume that the citations with `comparison' tag are the baselines used in the paper.
Dong and Schäfer~\cite{dong2011ensemble} classified citations into four categories i.e., `background', `fundamental idea', `technical basis' and `comparison'. They employed ensemble learning model for the classification. We also consider this as a relevant method for our task assuming that the citations tagged as `comparison' are the baselines of the paper.
Chakraborty and Narayanam~\cite{chakraborty2016all} measured how relevant a citation is w.r.t the citing paper and assigned five granular levels to the citations. Citations with level-5 are those which are extremely relevant and occur multiple times within the citing paper. We treat this work as another competing method for the current paper by considering citations tagged with level-5 as the baselines of the citing paper.
Jurgens et al.~\cite{jurgens2018measuring} built a classifier to categorize citations based on their functions in the text. The `comparison or contrast' category expresses the similarity/differences to the cited paper. This category might include some citations which are not considered for direct comparison, but they are the closest category to be considered as baseline. However, we have not compared with this method as as the proposed approach by Cohan et al.~\cite{cohan2019structural}, which is a baseline for the current work, already claimed to achieve better performance than this classifier.
Su et al.~\cite{su2019neural} used a single-layer convolutional neural network to classify citations and showed that it outperforms state-of-the-art methods. We also consider this as a baseline for our work.
Cohan et al.~\cite{cohan2019structural}  used a multi-task learning framework (using BiLSTM and Attention) and outperformed the approach of Jurgens et al.~\cite{jurgens2018measuring} on the citation classification task. Their `results comparison' category can be thought of as equivalent to the baseline class. This model achieved state-of-the-art performance on citation classification and we consider it as another baseline for our work.

\paragraph{\textbf{Recommending Citations for Scholarly Articles.}} A survey presented by Beel et al.~\cite{beel2016paper} showed that among ~200 research articles dealing with citation recommendation, more than half used content-based filtering on authors, citations and topics of the paper. Few such models include  topic-based citation recommendation~\cite{tang2009discriminative} and content-based recommendation~\cite{bhagavatula2018content,ding2014content} that work even when the metadata information about the paper being queried is missing. Yang et al.~\cite{yang2018lstm} used the LSTM model to develop a context-aware citation recommendation system. Recently, Jeong et al.~\cite{jeong2019context} developed a context-aware neural citation recommendation model. While there are a lot of new methods coming in the domain of citation recommendation systems, the problem of identifying and recommending baselines of a paper has been untouched. Citation recommendation can help researchers to efficiently write a scientific article, while baseline recommendation can further enable to get a glance at the work done in a particular domain.

\section{Dataset for Baseline Classification}
\label{sec:dataset}
We used ACL Anthology Reference Corpus (ARC)~\cite{aclanthology} as the base data source for preparing the annotated dataset for our study. The ARC corpus consists of scholarly papers published at various Computational Linguistics up to December 2015. The corpus consists of $22,875$ articles and provides the original PDFs, extracted text and logical document structure (section information) of the papers, and parsed citations using the ParsCit tool~\cite{parscit}. 
    
The complete ARC corpus contains all types of papers presented at various conferences under the ACL banner such as long and short research papers, system and demonstration papers, workshop and symposium papers. We noted that a significant fraction of short and workshop papers, and system and demonstration papers are not useful for our purpose as these papers often do not contain rigorous comparative evaluation. They generally are position papers, describe tools/systems, or  work in progress. Therefore, we discarded such articles from the dataset by removing papers having keywords such as \emph{short papers, workshops, demo, tutorial, poster, project notes, shared task,  doctoral consortium, companion volume}, and \emph{interactive presentation} in the title/venue fields of the papers. This filtering resulted in a final set of $8,068$ papers. 

We recruited two annotators, $A_1$ and $A_2$, for annotating the references of papers as baseline references. $A_1$ was a senior year undergraduate student, and $A2$ was a graduate student. Both the annotators were from the Computer Science discipline and had a good command of the English language (English being the primary medium of education).
\begin{wraptable}{r}{0.6\textwidth}
 \vspace{-10mm}
 \caption{Summary of the annotated dataset. Annotators $A1$ and $A2$ provided annotations for a total of $1,200$ and $1,000$ papers, respectively.}
\centering
\resizebox{0.6\textwidth}{!}{
    \begin{tabular}{@{}lccc@{}}
    \toprule
    & \begin{tabular}[c]{@{}l@{}}\# Papers\end{tabular} & \begin{tabular}[c]{@{}l@{}}\# Baseline\\ references\end{tabular} & \begin{tabular}[c]{@{}l@{}}\# Non-baseline\\ references\end{tabular} \\
    \midrule
    \textbf{Annotator 1} (A1) & 1,200 & 3,048 & 29,474 \\
    \textbf{Annotator 2} (A2) & 1,000 & 2,246 & 24,831 \\
    \textbf{Common Papers} & 125 & 305 & 3,252  \\
    \textbf{Unique Papers} & 2,075 & 4,989 & 51,053 \\
  \bottomrule
\end{tabular}
}
\vspace{-10mm}
  \label{tab:dataset_info}
\end{wraptable}
$A_1$ provided annotations for a total of $1,200$ documents selected randomly from the filtered list of $8,068$ papers. $A_2$ worked independently of $A_1$ and provided annotations for a total of $1,000$ papers. The set of documents annotated by $A_2$ had $875$ randomly selected new documents from the filtered ARC corpus and $125$ documents chosen randomly from the documents annotated by $A_1$. We used this set of $125$ papers annotated by both $A_1$ and $A_2$ to measure the inter-annotator agreement between them. The value of Cohen’s Kappa was found to be $0.913$ indicating near-perfect agreement between the two annotators.

We now discuss some of the challenges faced and observations made by the annotators while examining the assigned papers. The annotators noted that there were no associated citations for the baseline methods in the paper in many cases. This often happens when a well-established technique (such as tf-idf for document retrieval) or a simple method (such as a majority class baseline, a random classifier, a heuristic as a baseline) is used as a baseline.  Second, there were cases where the authors reported that it was difficult for them to compare their methods with other published techniques due to the novelty of the problem making published techniques unsuitable for their task. Finally, there were many cases where ideas from multiple papers were combined to create a suitable baseline for the task considered, making it hard and challenging to identify the baseline reference.

Table~\ref{tab:dataset_info} summarizes the statistics of the annotated dataset. The final dataset consists of $2,075$ unique papers. These papers have a total of $56,052$ references, out of which $4,989$ references were marked as baselines, and the remaining $51,053$ references were non-baseline references. 

\begin{table}[t!]
\caption{Distribution of papers in the dataset across different time periods.}
\centering
\setlength{\tabcolsep}{2.5pt}
\begin{small}
\begin{tabular}{@{}lrrrr@{}}
\toprule
 & \begin{tabular}[c]{@{}l@{}}1980-2000\end{tabular} & \begin{tabular}[c]{@{}l@{}}2001-2005\end{tabular} & \begin{tabular}[c]{@{}l@{}}2006-2010\end{tabular} & \begin{tabular}[c]{@{}l@{}}2011-2015\end{tabular} \\ \midrule
\textbf{\# Papers} & 125 & 179 & 589 & 1,182 \\
\textbf{\# References} & 2,339 & 3,534 & 13,976 & 36,193 \\
\textbf{\# Baselines} & 192 & 406 & 1,295 & 3.096 \\
\textbf{\begin{tabular}[c]{@{}l@{}}Mean references per paper\end{tabular}} & 18.71 & 19.74 & 23.73 & 30.62 \\
\textbf{\begin{tabular}[c]{@{}l@{}}Mean baselines per paper\end{tabular}} & 1.53 & 2.27 & 2.20 & 2.62 \\ \bottomrule
\end{tabular}%
\end{small}
\label{tab:year_wise_paper_distribution}
\end{table}

\subsection{Observations and Characteristics of the Dataset}
 
\textbf{Year-wise Distribution of Annotated Papers:} Table~\ref{tab:year_wise_paper_distribution} presents the year-wise distribution of the $2,075$ papers in the final dataset. The oldest paper in the dataset is from $1980$, and the latest paper is from $2015$. Table~\ref{tab:year_wise_paper_distribution} shows that papers published in the period $2011-2015$ cite more papers and have more baselines on an average compared to the papers published in the earlier years. This observation is consistent with the trend of an increased number of citations in papers~\cite{increasing-citations} and the increased focus on empirical rigor and reproducibility.

\begin{table}[t!]
    \caption{List of keywords used to identify the five section categories.}
\centering
\begin{small}
    \begin{tabular}{@{}p{0.3\columnwidth}p{0.65\columnwidth}@{}}
    \toprule
    \textbf{Section Heading}  & \textbf{Keywords} \\
    \midrule
    \textit{Introduction} & introduction \\
    \textit{Related Work} & related work; background; previous work; study \\
    \textit{Methods and Results} & method; approach; architect; experiment; empiric; evaluat; result; analys; compar; perform; discussion \\
    \textit{Conclusion} & conclusion; future work \\
    \textit{Other sections} &  everything else \\
    \bottomrule
    \end{tabular}
    \end{small}
\label{tab:section_description}
\end{table} 

\noindent
\textbf{Section-wise Distribution of Baseline Citations:} We now present the distribution of baseline references in different sections of papers in the dataset. Due to the diversity of writing styles and author preferences, there are no standardized section headers that are used in literature, and it is common to use simple rules, regular expressions~\cite{ding2013distribution}, or simple feature-based classification methods~\cite{algorithmseer} to identify section headers from document text. We use a simple keyword-based approach to group all the sections into five categories -- Introduction, Related Work, Methods and Results, Conclusions, and Others. A section of a paper containing a keyword as specified in Table~\ref{tab:section_description} would be mapped to its corresponding section category.

Table~\ref{tab:baseline_distribution} reports the distribution of baseline citations in different sections of the papers in our dataset. Note that a paper can be cited multiple times in the citing paper. Thus, a given citation can occur in multiple sections in a paper. We provide both the statistics, i.e., the total number of baseline citations in a section and the number of baseline citations that appear exclusively in the section in parenthesis.
\begin{wraptable}{r}{0.5\textwidth}
  \vspace{-7mm}
  \caption{Distribution of baselines and  non-baselines in different sections. Numbers in parentheses are the count of baselines appearing exclusively in the section.}
 \centering
 \begin{small}
  \begin{tabular}{@{}lll@{}}
        \toprule
        \textbf{Section} & \textbf{\# baselines} & \textbf{\#non-baselines}\\
      \midrule
      \textit{Introduction} & 2,138 (117) & 13,930 (7,360) \\
      \textit{Related} & 1,755 (105) & 14,917  (9,217)  \\ 
      \textit{Experiment} & 3,664 (534) & 11,939  (6,173)  \\ 
      \textit{Conclusion} & 203 (3) &     873  (360) \\   
      \textit{Other Sections} & 1,769 (181) & 13,283  (7,646)\\ 
    \bottomrule
   \end{tabular}
   \end{small}
   \vspace{-8mm}
    \label{tab:baseline_distribution}
\end{wraptable}

Interestingly, we note that there are a few cases where the baseline citations appear exclusively in the Introduction ($117$) and Conclusion ($3$) sections. One would expect the baseline citations not to appear exclusively in these sections. However, it turned out that the citations occurring exclusively in the conclusion section were part of a comparison table placed at the end of the paper. Therefore, they were counted under the conclusion section.  Further, the citations in the Introduction and Related Work section were given an alias name when they were first mentioned in the paper (e.g. LocLDA for location based LDA, see Table~\ref{tab:errors} for example) and were referred to by the aliases in other sections. Therefore, their presence in other sections of the paper could not be easily counted.
\begin{wraptable}{r}{0.5\textwidth}
  \vspace{-10mm}
\caption{Precision and recall values obtained by a na\"ive classifier that considers all citations in a specific section or table as baseline citations.}
    \centering
    \begin{small}
      \begin{tabular}{@{}lcc@{}}
        \toprule
          \textbf{Section Heading} & \textbf{Precision} & \textbf{Recall} \\
          \midrule
          \emph{Introduction} &  0.13 & 0.42  \\
          \emph{Related} & 0.10 & 0.35 \\
          \emph{Experiment} & 0.234 & 0.734 \\
          \emph{Conclusion} & 0.18 & 0.040 \\
          \emph{Other Sections} & 0.11 & 0.35\\
          \emph{Table} &0.72  & 0.18 \\
      \bottomrule
    \end{tabular}
    \end{small}
    \vspace{-8mm}
      \label{tab:section-wise-pr}
\end{wraptable}

From Table \ref{tab:baseline_distribution}, we observe that most of the baseline citations appear in the experiment section. Therefore, classifying a reference as a baseline if it occurs in the experiment section may be considered as a naive solution and a very simple baseline. In Table~\ref{tab:section-wise-pr}, we present the results obtained by hypothetical classifiers that classify all the citations in a given section as a baseline. Note that we also report numbers for a classifier that considers all citations appearing in a Table as baselines.

We note that while such a simple classifier will be able to recover a large number of baselines from the Experiment section (high recall value of $0.734$), it will miss out on about $30\%$ baselines and will suffer from a very high number of false positives (very low precision of $0.234$). An opposite trend can be observed in the case of Tables -- most citations in Tables are baseline references (high precision of $0.72$); however, due to a very low recall ($0.18$), most of the baselines are missed by this simple classifier.

\section{Multi-Module Attention Based Baseline Classifier}
\label{sec:approach}

\begin{figure*}[t!]
    \centering
    \includegraphics[width=0.90\textwidth]{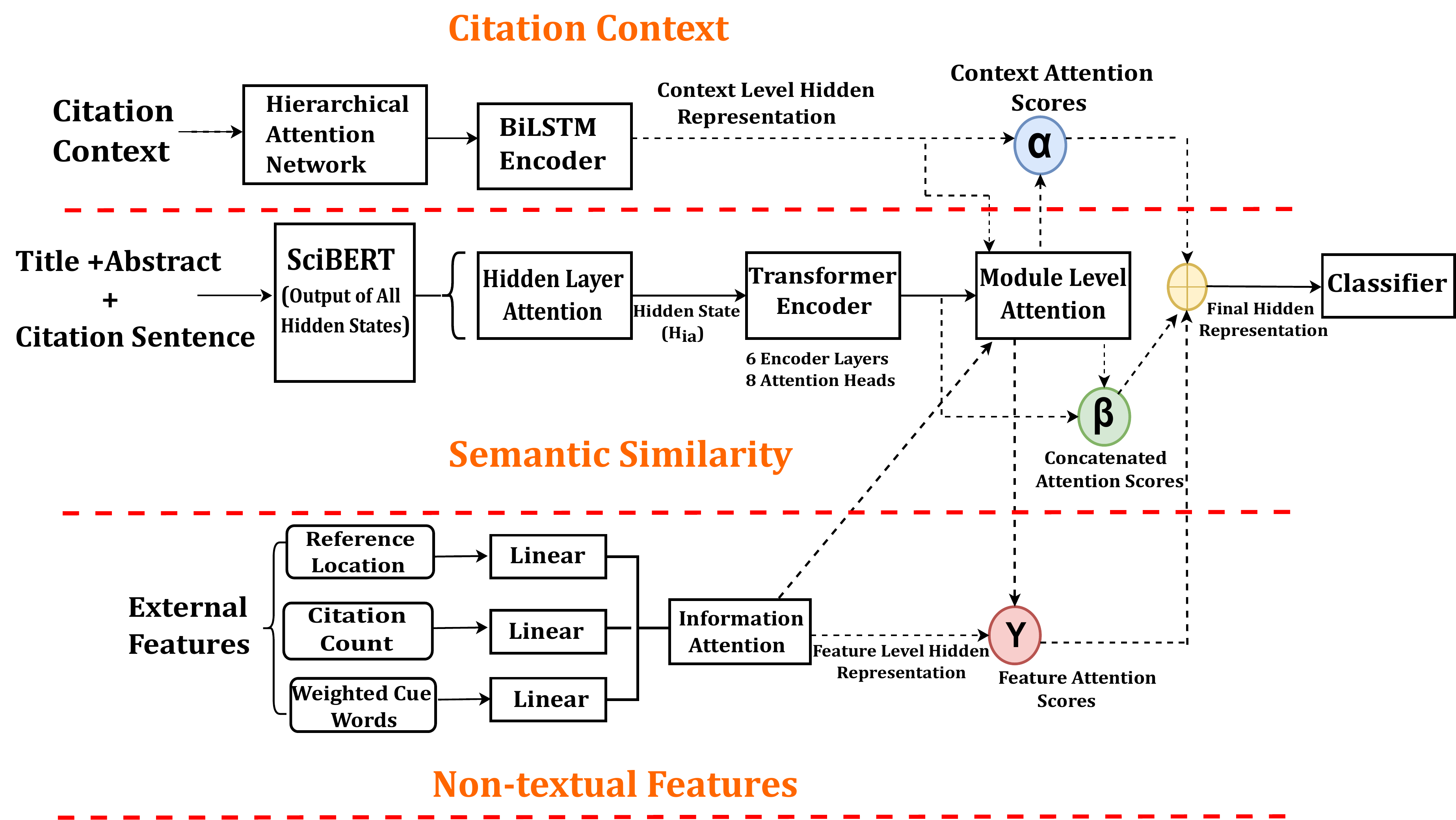} 
    \caption{Our proposed multi-module attention based neural classification model for the baseline classification task.}
    \label{fig:architecture}
\end{figure*}

We now describe our approach for classifying the citations of a paper as baselines. Our model utilizes contextual and textual signals present in the text around a citation to classify it as a baseline. We use Transformer encodings~\citep{transformer} to capture the nuances of the language and uses neural attention mechanisms ~\citep{transformer, yang-etal-2016-hierarchical} to learn to identify key sentences and words in the citation context of a citation. Further, given the vagaries of the natural language and varied writing styles of different authors, we also utilize non-textual signals such as popularity of a paper (in terms of its overall citations) to have a more robust classifier.

Fig.~\ref{fig:architecture} describes our proposed neural architecture for the baseline classification task. The proposed architecture is designed to capture different context signals in which a paper is cited to learn to differentiate between baselines and non-baseline citations. The proposed model utilizes a Transformer-based architecture consisting of three modules to handle different signals and uses the representations obtained from these modules together to classify a citation into a baseline.

The first module (top row in Fig.~\ref{fig:architecture}) tries to capture the intuition that the context around a citation in the paper can help in determining if the cited paper is being used as a baseline or not. Therefore, we take a fixed size context window and pass it through a hierarchical attention network~\cite{yang-etal-2016-hierarchical} that learns to identify and focus on sentences in the context window that can provide contextual clues about the cited paper being a baseline or not. Note that while selecting the context window, we ensure that all the sentences lie in the same paragraph as the citation under consideration. We select the size of the context window to be 10 sentences and for each sentence in the context window, we consider the sentence length to be 50 tokens. In case there are fewer sentences in a paragraph, we apply padding to ensure that the input to the network is of the same length. Similarly, we apply padding or pruning if the individual sentences are shorter or longer, respectively than 50 tokens. The citation context window thus obtained is then converted to a vector representation using SciBERT embeddings~\cite{Beltagy2019SciBERT} that provide word embeddings trained specifically for NLP applications using scholarly data.

The input vector representations thus obtained are fed to the hierarchical attention based encoder that outputs the hidden model representation of the context window after applying a series of localized attentions to learn the significance of constituent sentences and words in the input context vector. We show an example of the sentence level attention in Fig.~\ref{fig:h-attn}. The sentence containing the baseline citation (the middle sentence of the document) obtains the highest attention scores with rest of the attention distributed towards the other important sentences in the paragraph. This finally produces a better semantic understanding for the model in order to correctly classify it as a baseline.  The output of the hierarchical attention encoder model is then passed through a bidirectional LSTM encoder in order to capture any sequential relationships present in the citation context. This yields the final learned representation of the context surrounding the citation under consideration.

The second module (middle row in Fig.~\ref{fig:architecture}) is designed to capture the semantic similarity and relations between a given citation and the overall content of the citing paper. We consider the title and abstract of the citing paper as a concise summary of the citing paper. For a given citation, we take the title and abstract of the citing paper and the citation sentence and pass them through the pre-trained SciBERT language model that outputs a fine-tuned representation for the concatenated text. Further, we consider all the output hidden states for all the thirteen hidden layers in SciBERT. Different layers learn different feature representations of the input text. These representations from all the hidden layers, thus obtained are then passed through an attention module that learns attention weights for different hidden states. The resulting attention-weighted representation is then passed through a Transformer encoder layer\footnote{We use a six layer Transformer encoder with eight attention heads. This was found to be the best performing configuration.} to capture any sequential dependencies between input tokens yielding the final representation capturing relations between the cited paper and the title and abstract of the citing paper.

\begin{table*}[t!]
\caption{Cue words (after stemming) from the baseline contexts.}
    \centering
        \begin{tabular}{@{}p{.95\textwidth}@{}}
        \toprule
        \texttt{among base origin precis modifi highest implement extend signific maximum metric higher experi baselin fscore strategi accord compar overal perform best previou model evalu correl recal result calcul standard stateoftheart achiev figur accuraci gold comparison method top yield procedur obtain outperform score significantli increas report}\\
        \bottomrule
        \end{tabular}
      \label{tab:cuewords}
\vspace{-8mm}
\end{table*}

Note that the two modules discussed so far can capture the linguistic variations in the citation context and semantic relations between the cited and citing papers. In the third module (bottom row in Fig.~\ref{fig:architecture}), we utilize the following three additional non-textual signals that might indicate whether a paper is being cited as a baseline.

\begin{enumerate}
\item{\textbf{Reference location:}} Intuitively, if a paper is used as a baseline, it is more likely to be discussed (and cited) in the experiment section of the paper. Hence, we define five features that record the number of times a given reference is cited in each of the five sections defined in Table~\ref{tab:section_description}. In addition, we also define a feature to capture if a reference is cited in one of the tables as many times, baseline papers are also (and often exclusively) mentioned in the result-related tables.

\item{\textbf{Cue words:}} There are certain cue words and phrases that authors frequently use while discussing the baseline methods. Thus, their presence (or absence) in citation contexts can help differentiate between baseline and non-baseline references. We create a list of such cue words (as shown in Table~\ref{tab:cuewords}) by manually inspecting the citation contexts of baseline references in $50$ papers (separate from the papers in the dataset).  Thus, the cue word features capture the presence (or absence) of each cue word in the citation context of a reference. Further, each cue word $w$ present in the citation context is assigned a weight $w = 1/d_w$, where $d_w$ is the number of words between $w$ and the citation mention. Thus, cue words that appear near the citation mention are given a higher weight. If a cue word appear multiple times in the citation context, we consider its nearest occurrence to the citation mention (maximum weight).

\item{\textbf{Citation count:}} We use the total number of citations received by a paper as a feature to capture the intuition that highly-cited (and hence, more popular and impactful) papers have a higher chance of being used as a baseline than papers with low citations.
\end{enumerate}

Each of these features is then passed through a linear layer followed by a feature level attention module that yields the final attention weighted representation of all the features.

The output of the three modules described above provides three different representations capturing different information signals that can help the network classify the given citation as  baseline. The three representations thus obtained are passed through a module-level attention unit that learns attention weights to be given to the output of the three representations and outputs a 128 dimensional attention-weighted representation which is then passed through a linear classifier that outputs if the input citation is a baseline citation or not.

\section{Empirical Results and Discussions}
\label{sec:experiments}

\subsubsection{Baselines for Citation Classification:}
We select following methods for citation classification and adopt them for the task of baseline classification. We use author provided source-code where available; otherwise, we implement the methods using details and parameter settings as provided in the respective papers.

\begin{enumerate}

\item Dong and Schäfer~\cite{dong2011ensemble} proposed an ensemble-style self-training classifier to classify the citations of a paper into four categories -- \textit{background}, \textit{fundamental idea}, \textit{technical basis} and \textit{comparison}. We implemented their classifier (using their feature set) and used it for baseline classification task.

\item Chakraborty and Narayanam~\cite{chakraborty2016all} proposed a method for measuring relevance of a citation to the citing paper on a five point scale with level-5 citations being the most relevant. We consider the citations identified as level-5 as the baselines of the citing paper.

\item Su et al.~\cite{su2019neural} proposed a CNN based architecture for citation function classification that we use for our binary classification task.

\item Cohan et al.~\cite{cohan2019structural} proposed a multi-task learning framework for the citation classification task. We implement the model using the settings as recommended in the paper and use it for baseline classification.
\end{enumerate}
\comment{
    \item Simple heuristics \\
    Table~\ref{tab:pr_results_test} shows the precision and recall values of different sections on the test data. We observe the pattern is similar to what we observed on complete dataset. 
    \begin{table}[hbt!]
    \centering
      \caption{Section wise precision and recall values on test dataset}
      \label{tab:pr_results_test}
      \begin{tabular}{@{}lcc@{}}
        \toprule
          \textbf{Section Heading} & 
          \textbf{Precision} & 
          \textbf{Recall} \\
          \midrule
          \textbf{Experiment} & 0.23  & 0.79 \\
          \textbf{Conclusion} & 0.20 & 0.04\\
          \textbf{Introduction} & 0.13  &  0.47 \\
          \textbf{Other Sections} & 0.11 & 0.39\\
          \textbf{Related} & 0.09 & 0.37 \\ 
          \textbf{Table} & 0.74 & 0.20\\
      \bottomrule
    \end{tabular}
    \end{table}
}

\begin{table*}[!t]
  \centering
    \caption{Performance on baseline classification task for the different methods. We report overall precision, recall, and F-1 values as well as the numbers for each class.}
 \resizebox{0.99\textwidth}{!}{
 \begin{tabular}{@{}l @{~~}ccc@{~~} @{~~}ccc@{~~} @{~~}ccc@{}}
  \toprule
\multirow{2}{*}{\textbf{Models}} & \multicolumn{3}{c}{\textbf{Baselines}} & \multicolumn{3}{c}{\textbf{Non-baselines}} & \multicolumn{3}{c}{\textbf{Overall}} \\
\cmidrule(lr){2-4} \cmidrule(lr){5-7} \cmidrule(lr){8-10}
 & \textbf{Precision} & \textbf{Recall} & \textbf{F-1} & \textbf{precision} & \textbf{Recall} & \textbf{F-1} & \textbf{Precision} & \textbf{Recall} & \textbf{F-1} \\
    \midrule
    Dong and Schäfer~\cite{dong2011ensemble} & 0.33 & 0.67 & 0.44 & 0.96 & 0.87 & 0.91 & 0.65 & 0.77 & 0.68 \\
    Chakraborty and Narayanam~\cite{chakraborty2016all} & 0.26 & \textbf{0.74} & 0.39 & 0.96 & 0.78 & 0.86 & 0.61 & 0.76 & 0.62 \\
    Su et al.~\cite{su2019neural} & 0.69  & 0.16 & 0.26 & 0.63 & 0.95 &  0.76 & 0.66 & 0.55 & 0.51 \\
    Cohan et al.~\cite{cohan2019structural} & 0.47 & 0.48 & 0.47 & 0.96 & 0.95 & 0.95 & 0.71 & 0.71 & 0.71 \\
    \midrule
    Proposed MMA classifier & \textbf{0.69} & 0.57 & \textbf{0.63} & \textbf{0.96} & \textbf{0.98} & \textbf{0.97} & \textbf{0.82} & \textbf{0.78} & \textbf{0.80} \\ 
    \bottomrule
\end{tabular}
}
  \label{tab:reults}
 \vspace{-5mm}
\end{table*}

\subsubsection{Experimental Settings:} For evaluating different classification methods, we split the developed dataset (Section~\ref{sec:dataset}) into training, development, and test sets in $70:10:20$ ratio. Different hyper-parameters involved are fine-tuned using the development set. Consequently, the size of the input citation context vectors is set to $768$, the size of the hidden layer for the BiLSTM layer is $64$ and the dropout rate is set to $0.2$. The Transformer encoder has 6 layers and 8 attention heads. The batch size and learning rate are set to $32$ and $ 0.001$, respectively. The model was trained for $20$ epochs. For our proposed model, we used cross-entropy loss and Adam Optimizer~\cite{adam} to minimize the overall loss of the model. As our dataset is unbalanced, we incorporated class weights in our loss function fine-tuned the class weights. 

\subsubsection{Results and Discussions:} Table~\ref{tab:reults} summarizes the results as achieved by different methods on the test set. We note that four state-of-the-art methods for citation classification achieve only moderate performance on the baseline classification task indicating their inadequacy at this task, and hence, the need for developing specialized methods for baseline classification. Our proposed model, outperforms the state-of-the-art citation role classifiers in terms of F-1 measure. Further, note that the performance of the proposed Multi-module Attention based model is more balanced with relatively high recall ($0.57$) and the highest precision($0.69$) among all the methods studied.

\begin{figure*}[!t]
    \centering
    \includegraphics[width=\textwidth]{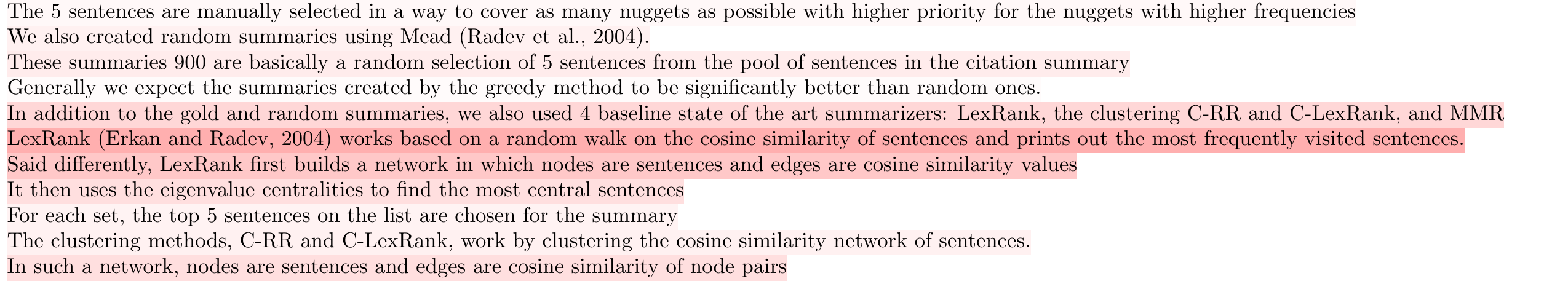} 
    \caption{ Example of a sentence-level attention distribution (Red) obtained from the Attention Encoder.}
    \label{fig:h-attn}
\vspace{-3mm}
\end{figure*}
\begin{figure*}[!t]
    \centering
    \includegraphics[width=\textwidth]{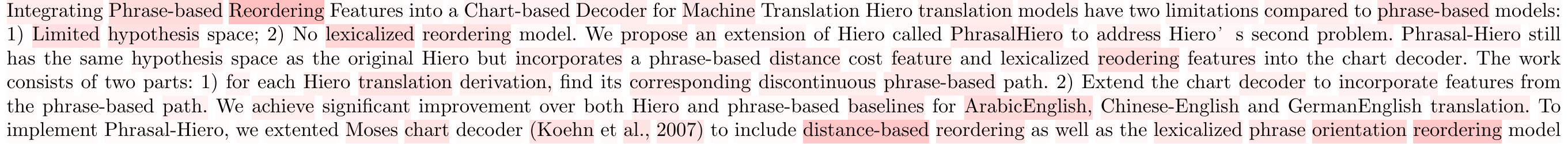} 
    \caption{ Illustrative example of an attention weight distribution (red) from the Attention Encoder in the semantic similarity module of the proposed network.}
    \label{fig:s-attn}
\vspace{-3mm}
\end{figure*}

Fig.~\ref{fig:h-attn} shows an illustrative example of the hierarchical attention module in the proposed network. The figure shows the citation context as extracted from the paper by Qazvinian et al.~\cite{qazvinian-etal-2010-citation} where the LexRank method by Erkan and Radev~\cite{Erkan_2004} is being used as a baseline. The attention given to different sentences in the context window is illustrated by shades of red where a sentence in darker shade is given a higher weight. We note that the sentence which the LexRank paper is cited, is given the highest weight and other sentences that talk about the task of summarization are also given some weights whereas the fourth sentence (``Generally we expect...") is being given no weight as the network did not find it to be useful for the classification task. 
Likewise, Fig.~\ref{fig:s-attn} presents an example of the role of the attention encoder in the semantic similarity module in the proposed network. The figure shows the concatenated title and abstract of the paper by Nguyen and Vogel~\cite{s-attn} that uses the MOSES decoder~\cite{moses} for machine translation (last sentence in the figure is the citation sentence). Note that the network is able to identify keywords like \emph{reordering, distance-based, translation,} and \emph{lexicalized} that indicate the similarity between the content of the citing paper with the citation context.

\begin{table}[!t]
\centering
\caption{Example of false positives treated as baselines by the classifier. Paper IDs are the IDs used in the dataset.}
\label{tab:errors}
  \scalebox{0.7}{
  \begin{tabular}{@{}cp{\columnwidth}@{}}
    \toprule
      \textbf{Paper Id} & 
      \textbf{Citation text}  \\
      \midrule
      \textbf{N12-1051
            } & We evaluated our taxonomy induction algorithm using McRae et al.’s (2005) dataset which consists of for 541 basic level nouns.\\
        \textbf{P08-1027} 
        & For each parameter we have estimated its desired range using the (Nastase and Szpakowicz, 2003) set as a development set. \\
        \midrule
          \textbf{D13-1083 } & In the future work, we will compare structural SVM and c-MIRA under decomposable metrics like WER or SSER (Och and Ney, 2002).\\
        \textbf{E09-1027} & For comparison purposes, we plan to implement other features that have been used in earlier readability assessment systems. For example, Petersen and Ostendorf (2009) created lists of the most common words from the Weekly Reader articles,  \\
        \midrule
         \textbf{P10-1116  } 
                & This is in line with results obtained by previous systems (Griffiths et al., 2005; Boyd- Graber and Blei, 2008; Cai et al., 2007). While the performance on verbs can be increased to outperform the most frequent sense baseline. \\
        \midrule
          \textbf{D10-1006 } 
                & This is the model used in (Brody and Elhadad, 2010) to identify aspects, and we refer to this model as LocLDA.\\
        \textbf{D11-1115} & we compare Chart Inference to the two baseline methods: Brute Force (BF), derived from Watkinson and Manandhar, and Rule-Based (RB), derived from Yao et al.   \\
      \bottomrule
    \end{tabular}}
    \vspace{-7mm}
\end{table}

\subsubsection{Error Analysis:} We now present representative examples of hard cases and the types of errors made by the classifiers.
 
\textit{Confusion with Datasets:} We observed that often the citation for datasets used in the experiments were classified as baselines by the classifier. Such citations are often made in the experiment section,  and the language patterns in their citation contexts are often very similar to contexts of baseline citations (rows 1 and 2 in Table~\ref{tab:errors}).
 
\textit{Citations for Future Work:} Often, authors discuss the results of papers that are not explicitly used as baselines in the current work but are discussed for the sake of completeness and could be used as baselines as part of the future work. One could argue that such citations should be easy to classify as they must be part of the \emph{Conclusions and Future Work} sections. However, as we observed, this does not always hold true. Such citations could be found in the \emph{Experiment} or \emph{Other} custom section headers (e.g. rows 3, 4 in Table~\ref{tab:errors}).
 
\textit{Context Overlap of Multiple Citations:} The key assumption that the methods studied in this work make is that the baseline and non-baseline citations differ in the language patterns in their respective citation contexts. However, we noted that multiple papers are often cited together, and thus, share the same citation contexts (and other properties represented by different features).  For instance, row 5 in Table~\ref{tab:errors} presents an example of non-baseline citations sharing the context with baseline \textit{(Cai et al. 2007)}.

\textit{Citation Aliases and Table Citations:} Often, authors give an alias to a particular method (as shown in rows 6, 7 in Table~\ref{tab:errors}) and then use the alias to refer to that method in the rest of the paper. As a result, it becomes challenging to capture the context around the alias mentions in the text.  Further, many errors were made in cases where the baseline references are not cited and discussed extensively in the running text but are mentioned directly in the results table. Hence, we lose out on the context for such baseline citations. 
\section{Conclusions}
We introduced the task of identifying the papers that have been used as baselines in a given scientific article. We framed the task as a reference classification problem and developed a dataset out of ACL anthology corpus for the baseline classification task. We empirically evaluated four state-of-the-art methods for citation classification and found that they do not perform well for the current task. We then developed custom classifiers for the baseline classification task. While the proposed methods outperformed the state-of-the-art citation classification methods, there is still a significant performance gap that needs to be filled. We further presented error analysis illustrating the challenges and examples that the proposed systems found difficult to classify.

\section*{Acknowledgement}
T. Chakraborty would like to acknowledge the support of the Ramanujan Fellowship, and ihub-Anubhuti-iiitd Foundation set up under the NM-ICPS scheme of the Department of Science and Technology, and the Infosys Centre for AI at IIIT-Delhi.

\bibliographystyle{splncs04}
\bibliography{references}

\end{document}